\documentclass{article} 
\usepackage{iclr2020_conference,times}

\usepackage{amsmath,amsfonts,bm}

\def\eqref#1{equation~\ref{#1}}

\def\1{\bm{1}}

\DeclareMathAlphabet{\mathsfit}{\encodingdefault}{\sfdefault}{m}{sl}
\SetMathAlphabet{\mathsfit}{bold}{\encodingdefault}{\sfdefault}{bx}{n}

\usepackage{hyperref}
\usepackage{url}
\usepackage{amsfonts}       
\usepackage{nicefrac}       
\usepackage{microtype}      
\usepackage[pdftex]{graphicx}
\usepackage{subcaption}
\usepackage[shortlabels]{enumitem}

\title{Coherent Gradients: An Approach to\\Understanding Generalization\\in
Gradient Descent-based Optimization}

\author{Satrajit Chatterjee\\
Google AI\\
Mountain View, CA 94043, USA \\
\texttt{schatter@google.com} \\
}

\iclrfinalcopy 
\begin{document}

\maketitle

\begin{abstract}

An open question in the Deep Learning community is why neural networks
trained with Gradient Descent generalize well on real datasets even though
they are capable of fitting random data. We propose an approach to
answering this question based on a hypothesis about the dynamics of
gradient descent that we call {\em Coherent Gradients}: Gradients from similar
examples are similar and so the overall gradient is stronger in certain
directions where these reinforce each other. Thus changes to the network
parameters during training are biased towards those that (locally)
simultaneously benefit many examples when such similarity exists. We
support this hypothesis with heuristic arguments and perturbative
experiments and outline how this can explain several common empirical
observations about Deep Learning. Furthermore, our analysis is not just
descriptive, but prescriptive. It suggests a natural modification to
gradient descent that can greatly reduce overfitting.

\end{abstract}

\section{Introduction and Overview}

Neural networks used in practice often have sufficient effective capacity to
learn arbitrary maps from their inputs to their outputs. This is typically
demonstrated by training a classification network that achieves good test
accuracy on a real dataset $S$, on a modified version of $S$ (call it $S'$)
where the labels are randomized and observing that the training accuracy on
$S'$ is very high, though, of course, the test accuracy is no better than
chance~\citep{Zhang17}. This leads to an important open question in the Deep
Learning community~(\citet{Zhang17,Arpit17,Bartlett17,Kawaguchi17,Neyshabur18,Arora18,Belkin19,Rahaman19,Nagarajan19}, etc.): Among all maps that fit a real dataset, how does Gradient
Descent (GD) find one that generalizes well? This is the question we
address in this paper. 

We start by observing that this phenomenon is not limited to neural networks
trained with GD but also applies to Random Forests and Decision Trees.
However, there is no mystery with trees: A typical tree construction algorithm
splits the training set recursively into similar subsets based on input
features. If no similarity is found, eventually, each example is put into its
own leaf to achieve good training accuracy (but, of course, at the cost of poor
generalization). Thus, trees that achieve good accuracy on a randomized dataset
are much larger than those on a real dataset (e.g. ~\citet[Expt. 5]{Chatterjee19}).

Is it possible that something similar happens with GD? We believe so. The type of
randomized-label experiments described above show that if there are common
patterns to be found, then GD finds them. If not, it fits each example on a
case-by-case basis. The question then is, what is it about the dynamics of GD
that makes it possible to extract common patterns from the data? And what does
it mean for a pattern to be common?

Since the only change to the network parameters in GD comes from the
gradients, the mechanism to detect commonality amongst examples must be through
the gradients. We propose that this commonality detection can be explained as
follows:
\begin{enumerate}

    \item Gradients are {\em coherent}, i.e, similar examples (or parts of examples) have
        similar gradients (or similar components of gradients) and dissimilar
        examples have dissimilar gradients.

    \item Since the overall gradient is the sum of the per-example gradients,
        it is stronger in directions where the per-example gradients are
        similar and reinforce each other and weaker in other directions where
        they are different and do not add up.
 
    \item Since network parameters are updated proportionally to gradients,
        they change faster in the direction of stronger gradients. 

    \item Thus the changes to the network during training are biased towards
        those that simultaneously benefit many examples instead of a few (or
        one example). 

\end{enumerate}
For convenience, we refer to this as the Coherent Gradients hypothesis.
 
It is instructive to work through the proposed mechanism in the
context of a simple thought experiment. Consider a training set with two
examples $a$ and $b$. At some point in training, suppose the gradient of $a$,
$g_a$, can be decomposed into two orthogonal components $g_{a_1}$ and
$g_{a_2}$ of roughly equal magnitude, i.e., there are two, equally good, independent ways in which the network can better
fit $a$ (by using say two disjoint parts of the network). Likewise, for
$b$. Now, further suppose that one of the two ways is common to both $a$ and
$b$, i.e., say $g_{a_2} = g_{b_2} = g_{ab}$, whereas, the other two are example specific, i.e., $\langle g_{a_1}, g_{b_1} \rangle = 0$. Now, the overall gradient is 
$$g = g_a + g_b = g_{a_1} + 2\ g_{ab} + g_{b_1}.$$
Observe that the gradient is stronger in the direction that simultaneously
helps both examples and thus the corresponding parameter changes are bigger
than those those that only benefit only one example.\footnote{
    While the mechanism is easiest to see with full or large minibatches,
    we believe it holds even for small minibatches (though there one has to consider
    the bias in updates over time).}

It is important to emphasize that the notion of similarity used above (i.e.,
which examples are considered similar) is not a constant but changes in the
course of training as network parameters change. It starts from a mostly task
independent notion due to random initialization and is bootstrapped in the
course of training to be task dependent. We say ``mostly" because even with
random initialization, examples that are syntactically close are treated
similarly (e.g., two images differing in the intensities of some pixels as
opposed to two images where one is a translated version of the other).  

The relationship between strong gradients and generalization can also be
understood through the lens of algorithmic stability~\citep{Bousquet02}: strong gradient
directions are more stable since the presence or absence of a single example
does not impact them as much, as opposed to weak gradient directions which may
altogether disappear if a specific example is missing from the training set.
With this observation, we can reason inductively about the stability
of GD: since the initial values of the parameters do not depend on the
training data, the initial function mapping examples to their gradients is
stable. Now, if all parameter updates are due to strong gradient directions,
then stability is preserved. However, if some parameter updates are due to weak
gradient directions, then stability is diminished. Since stability (suitably formalized) is
equivalent to generalization~\citep{ShalevShwartz10}, this allows us to see how generalization may
degrade as training progresses. Based on this insight, we shall see later how a
simple modification to GD to suppress the weak gradient directions can
dramatically reduce overfitting.

In addition to providing insight into why GD generalizes in practice, we
believe that the Coherent Gradients hypothesis can help explain several other empirical
observations about deep learning in the literature:
\begin{enumerate}[(a)]

    \item Learning is slower with random labels than with real labels~\citep{Zhang17,Arpit17}
    \item Robustness to large amounts of label noise~\citep{Rolnick17}
    \item Early stopping leads to better generalization~\citep{Caruana00}
    \item Increasing capacity improves generalization~\citep{Caruana00, Neyshabur18}
    \item The existence of adversarial initialization schemes~\citep{Liu19}
    \item GD detects common patterns even when trained with random labels~\citep{Chatterjee19}

\end{enumerate}

A direct experimental verification of the Coherent Gradients hypothesis is
challenging since the notion of similarity between examples depends on the
parameters of the network and thus changes during training.
Our approach, therefore, is to design intervention experiments where we
establish a baseline and compare it against variants designed to test some aspect
or prediction of the theory.
As part of these experiments, we replicate the observations (a)--(c) in the
literature noted above, and analyze the corresponding explanations provided by
Coherent Gradients (\S\ref{sec:noise}), and outline for future work how 
(d)--(f) may be accounted for (\S\ref{sec:future}).

In this paper, we limit our study to simple baselines: vanilla Stochastic
Gradient Descent (SGD) on MNIST using fully connected networks.
We believe that this is a good starting point, since even in this simple
setting, with all frills eliminated (e.g., inductive bias from architecture or
explicit regularization, or a more sophisticated optimization procedure), we
are challenged to find a satisfactory explanation of why SGD generalizes well.
Furthermore, our prior is that the difference between weak and strong
directions is small at any one step of training, and therefore having a strong
learning signal as in the case of MNIST makes a direct analysis of gradients easier.
It also has the benefit of having a smaller carbon footprint and being easier
to reproduce.
Finally, based on preliminary experiments on other architectures and datasets
we are optimistic that the insights we get from studying this simple setup
apply more broadly.

\section{Effect of Reducing Similarity Between Examples}
\label{sec:noise}

Our first test of the Coherent Gradients hypothesis is to see what happens when we
reduce similarity between examples. Although, at any point during training, we
do not know which examples are similar, and which are different, we can (with
high probability) reduce the similarity among training examples simply by
injecting label noise. In other words, under any notion of similarity, adding
label noise to a dataset that has clean labels is likely to make similar
examples less similar.
Note that this perturbation does not reduce coherence since gradients still
depend on the examples. (To break coherence, we would have to make the
gradients independent of the training examples which would requiring perturbing
SGD itself and not just the dataset).

\subsection{Setup} 
\label{sec:setup}

For our baseline, we use the standard MNIST dataset of 60,000 training examples
and 10,000 test examples. Each example is a 28x28 pixel grayscale handwritten
digit along with a label (`0'--`9').
We train a fully connected network on this dataset. The network has one hidden
layer with 2048 ReLUs and an output layer with a 10-way softmax.
We initialize it with Xavier and train using vanilla SGD (i.e., no momentum)
using cross entropy loss with a constant learning rate of 0.1 and a minibatch
size of 100 for $10^5$ steps (i.e., about 170 epochs). We do not use any
explicit regularizers.

We perturb the baseline by modifying {\em only} the dataset and keeping all
other aspects of the architecture and learning algorithm fixed.
The dataset is modified by adding various amounts of noise (25\%, 50\%, 75\%,
and 100\%) to the labels of the training set (but not the test set).
This noise is added by taking, say in the case of 25\% label noise, 25\% of the
examples at random and randomly permuting their labels. Thus, when we add 25\%
label noise, we still expect about 75\% + 0.1 * 25\%, i.e., 77.5\% of the
examples to have unchanged (i.e. ``correct") labels which we call the {\em
proper accuracy} of the modified dataset. In what follows, we call examples
with unchanged labels, {\em pristine}, and the remaining, {\em corrupt}.
Also, from this perspective, it is convenient to refer to the original MNIST
dataset as having 0\% label noise.

We use a fully connected architecture instead of a convolutional one to
mitigate concerns that some of the difference in generalization between the
original MNIST and the noisy variants could stem from architectural inductive
bias.
We restrict ourselves to only 1 hidden layer to have the gradients be as
well-behaved as possible.
Finally, the network width, learning rate, and the number of training steps are
chosen to ensure that exactly the same procedure is usually able to fit all 5
variants to 100\% training accuracy.

\subsection{Qualitative Predictions} 
\label{sec:pred}

Before looking at the experimental results, it is useful to consider what
Coherent Gradients can qualitatively say about this setup. In going from 0\%
label noise to 100\% label noise, as per experiment design, we expect examples
in the training set to become more dissimilar (no matter what the current
notion of similarity is).  
Therefore, we expect the per-example gradients to be less aligned with each
other. This in turn causes the overall gradient to become more diffuse, i.e.,
stronger directions become relatively weaker, and consequently, we expect it to
take longer to reach a given level of accuracy as label noise increases, i.e., to have a 
lower {\em realized learning rate}.


This can be made more precise by considering the following heuristic argument.
Let $\theta_t$ be the vector of trainable parameters of the network at training step
$t$. Let $\mathcal{L}$ denote the loss function of the network (over all training examples).
Let $g_t$ be the gradient of $\mathcal{L}$ at $\theta_t$ and let $\alpha$ denote the
learning rate. By Taylor expansion, to first order, the change $\Delta
\mathcal{L}_t$ in the loss function due to a small gradient descent step $h_t =
-\alpha \cdot g_t$ is given by
\begin{equation}
    \label{eq:1}
    \Delta \mathcal{L}_t 
    := \mathcal{L}(\theta_t + h_t) - \mathcal{L}(\theta_t) 
    \approx \langle g_t, h_t \rangle
    = -\alpha \cdot \langle g_t, g_t \rangle
    = -\alpha \cdot \|g_t\|^2
\end{equation}
where $\|\cdot\|$ denotes the $l_2$-norm. Now, let $g_{te}$ denote the gradient
of training example $e$ at step $t$. Since the overall gradient is the sum of
the per-example gradients, we have,
\begin{equation}
    \label{eq:2}
    \|g_t\|^2 
    = \langle g_t, g_t \rangle
    = \langle \sum\limits_{e} g_{te}, \sum\limits_{e} g_{te} \rangle
    = \sum\limits_{e, e'} \langle g_{te}, g_{te'} \rangle
    = \sum\limits_{e} \| g_{te} \|^2 + \sum\limits_{\substack{e, e' \\ e \neq e'}} \langle g_{te}, g_{te'} \rangle
\end{equation}

Now, heuristically, let us assume that all the $\|g_{te}\|$ are roughly the same and equal to
$\|g_t^\circ\|$ which is not entirely unreasonable (at least at the start of
training, if the network has no {\em a priori} reason to treat different
examples very differently).
If all the per-example gradients are approximately orthogonal (i.e., $\langle
g_{te}, g_{te'} \rangle \approx 0$ for $e \neq e'$), then $\|g_t\|^2 \approx m
\cdot \|g_t^\circ\|^2$ where $m$ is the number of examples. On the other hand,
if they are approximately the same (i.e., $\langle g_{te}, g_{te'} \rangle
\approx \|g_t^\circ\|^2$), then $\|g_t\|^2 \approx m^2 \cdot \|g_t^\circ\|^2$.
Thus, we expect that greater the agreement in per-example gradients, the faster
loss should decrease.

Finally, for datasets that have a significant fractions of pristine and corrupt
examples (i.e., the 25\%, 50\%, and 75\% noise) we can make a more nuanced
prediction.
Since, in those datasets, the pristine examples as a group are still more
similar than the corrupt ones, we expect the pristine gradients to continue to
align well and sum up to a strong gradient.
Therefore, we expect them to be learned faster than the corrupt examples, and at
a rate closer to the realized learning rate in the 0\% label noise case.
Likewise, we expect the realized learning rate on the corrupt examples to be
closer to the 100\% label noise case.
Finally, as the proportion of pristine examples falls with increasing noise, we
expect the realized learning rate for pristine examples to degrade.

Note that this provides an explanation for the observation in the literature
that that networks can learn even when the number of examples with noisy labels
greatly outnumber the clean examples {\em as long as the number of clean
examples is sufficiently large}~\citep{Rolnick17} since with too few clean
examples the pristine gradients are not strong enough to dominate.

\subsection{Agreement with Experiment} 

\begin{figure}[t]
\begin{center}
\begin{subfigure}[b]{0.49\textwidth}
    \includegraphics[width=\textwidth]{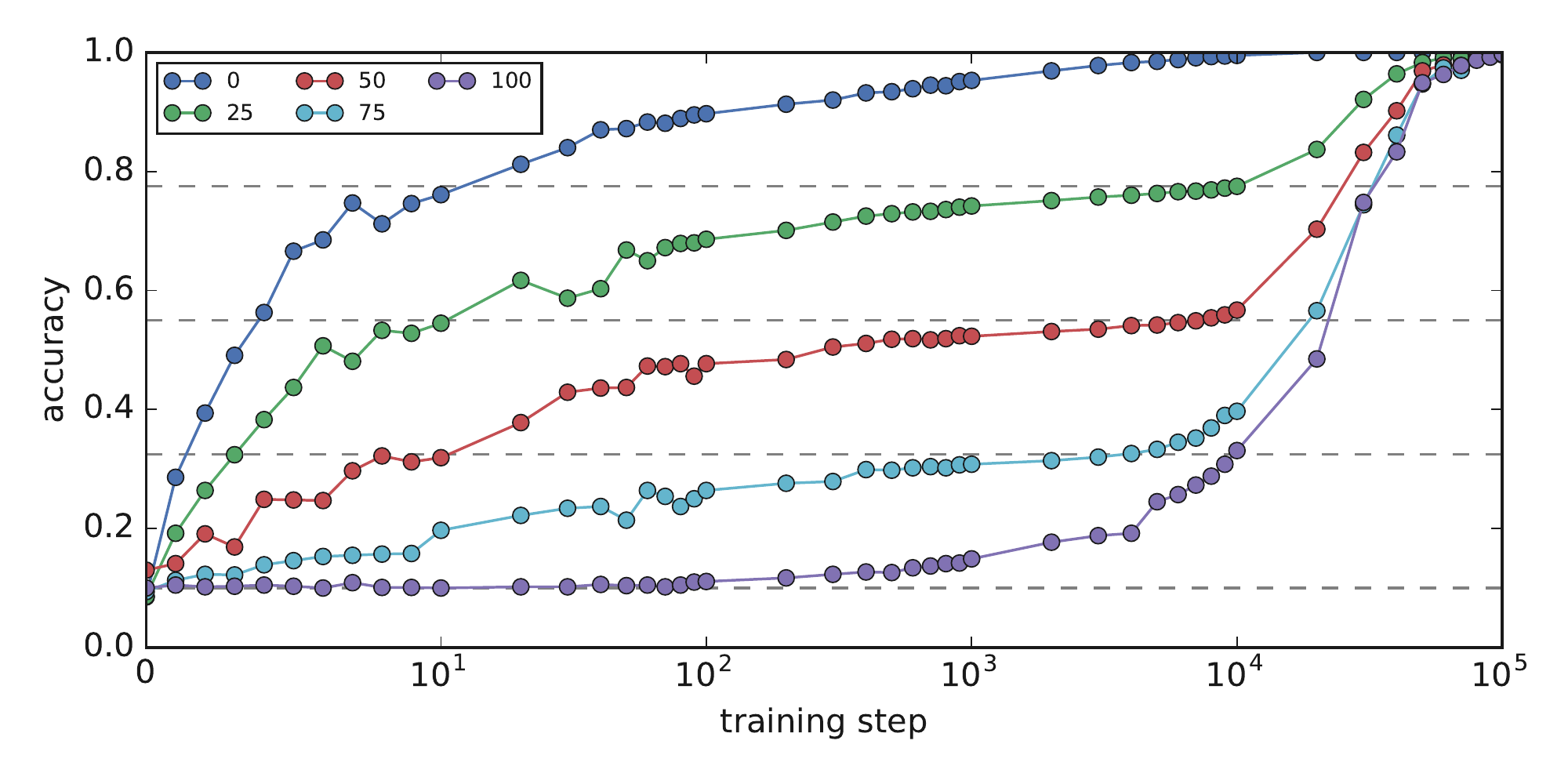}
    \caption{Training accuracy.}
\end{subfigure}
~ 
\begin{subfigure}[b]{0.49\textwidth}
    \includegraphics[width=\textwidth]{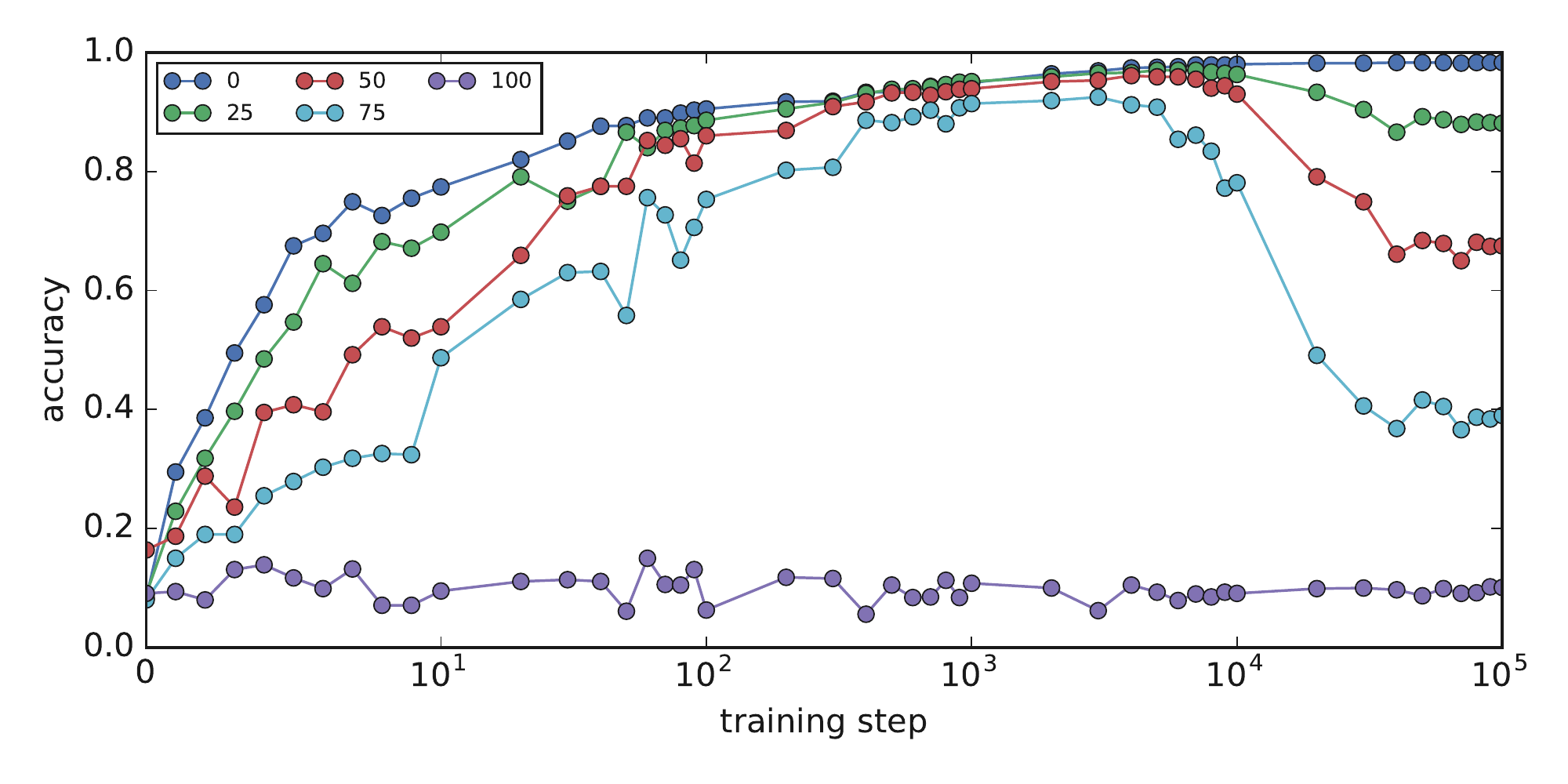}
    \caption{Validation accuracy.}
\end{subfigure}

\begin{subfigure}[b]{0.49\textwidth}
    \includegraphics[width=\textwidth]{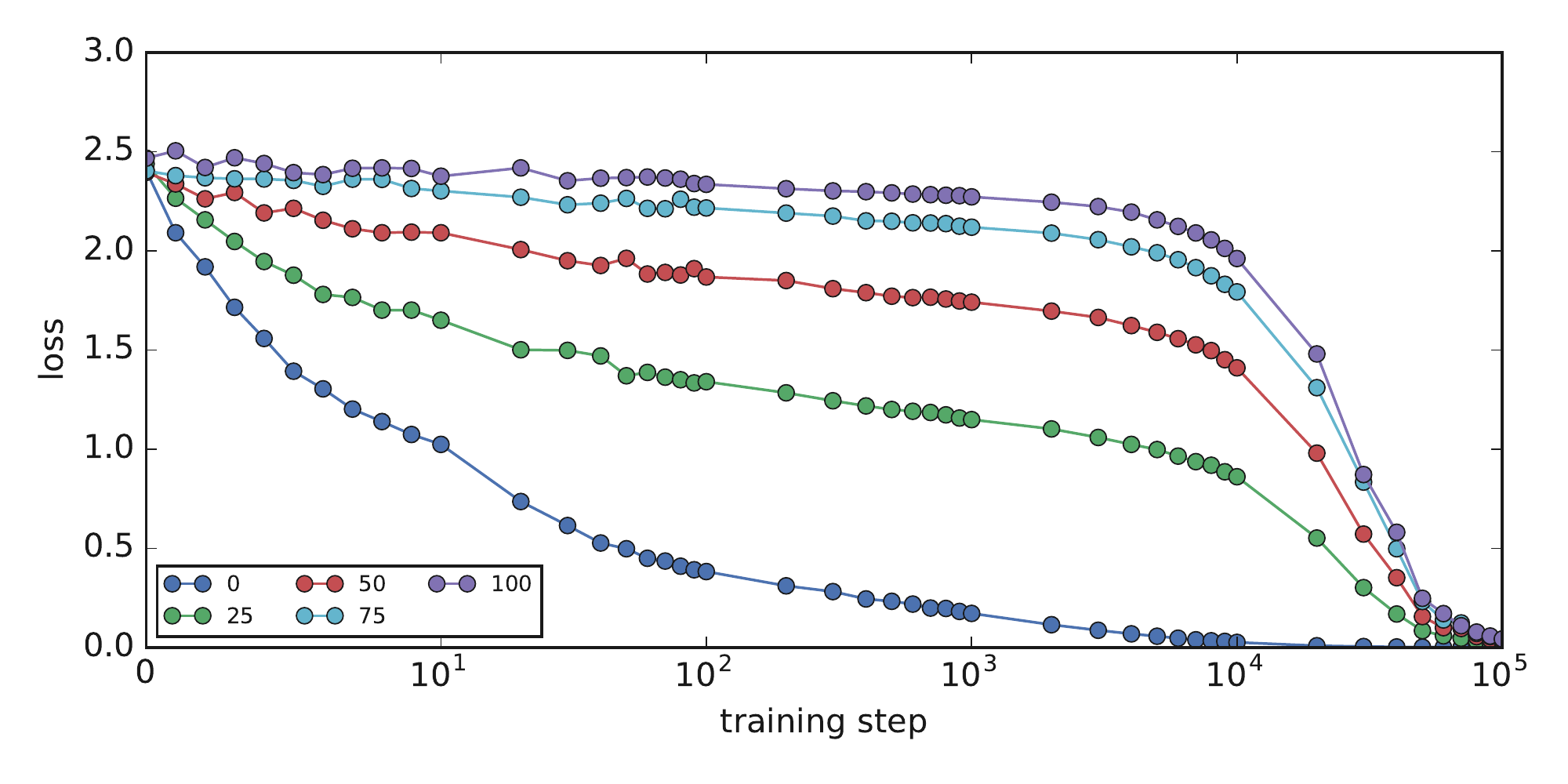}
    \caption{Training loss.}
\end{subfigure}
~
\begin{subfigure}[b]{0.49\textwidth}
    \includegraphics[width=\textwidth]{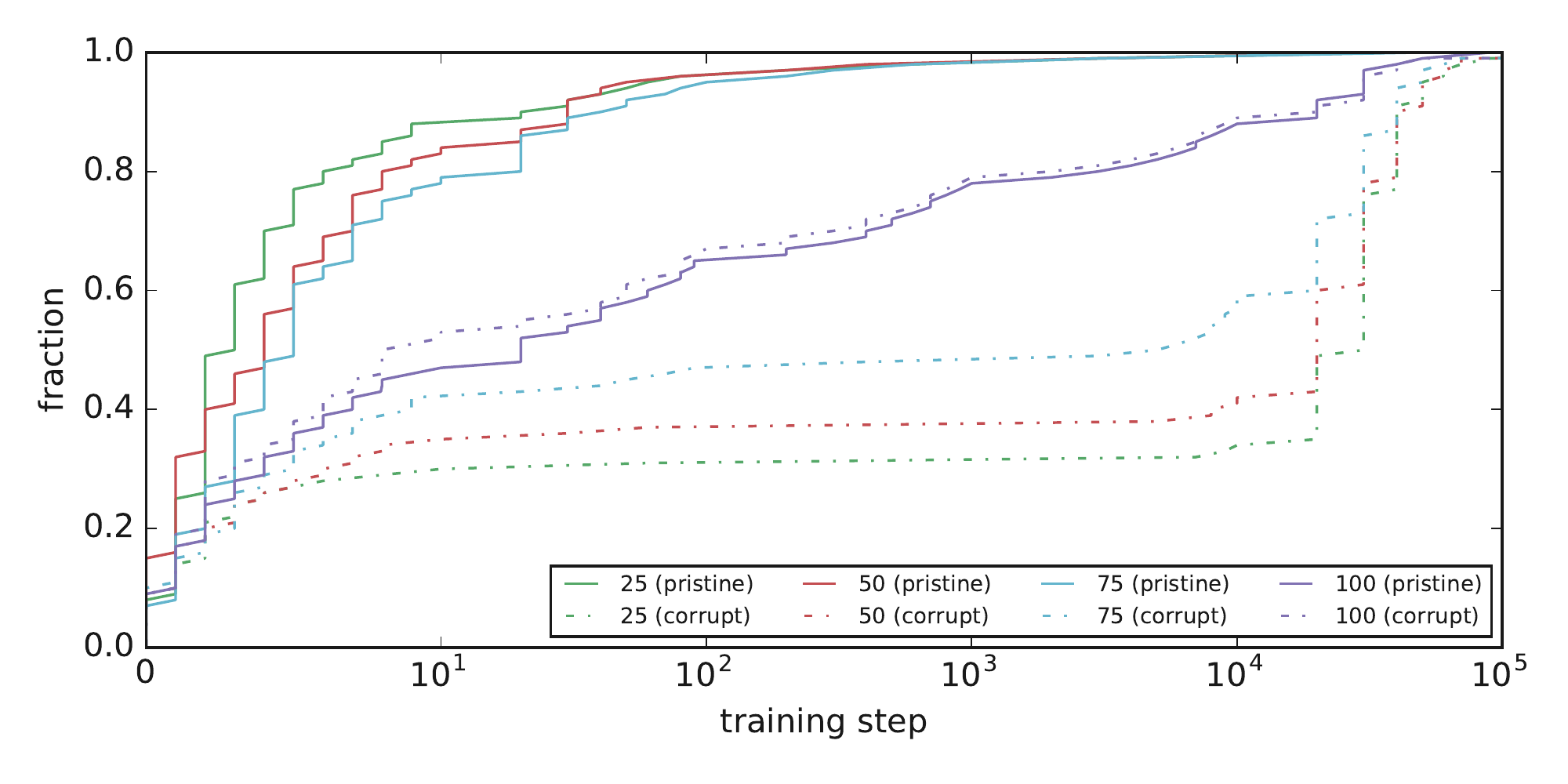}
    \caption{First training step when an example is learned.}
\end{subfigure}
\end{center}
    \caption{Results of the experiment to reduce similarity by adding label noise (\S\ref{sec:noise}).}
\label{fig:ta}
\end{figure}

Figure~\ref{fig:ta}(a) and (b) show the training and test curves for the baseline and the
4 variants. We note that for all 5 variants, at the end of training, we achieve
100\% training accuracy but different amounts of generalization. As expected,
SGD is able to fit random labels, yet when trained on real data, generalizes well.
Figure 1(c) shows the reduction in training loss over the course of training,
and Figure 1(d) shows the fraction of pristine and corrupt labels learned as
training processes.

The results are in agreement with the qualitative predictions made above:
\begin{enumerate}

    \item In general, as noise increases, the time taken to reach a given level
        of accuracy (i.e., realized learning rate) increases.

    \item Pristine examples are learned faster than corrupt examples. They
        are learned at a rate closer to the 0\% label noise rate whereas the corrupt
        examples are learned at a rate closer to the 100\% label noise rate.

    \item With fewer pristine examples, their learning rate reduces. This is
        most clearly seen in the first few steps of training by comparing say
        0\% noise with 25\% noise.

\end{enumerate}

Using Equation~\ref{eq:1}, note that the magnitude of the slope of the training
loss curve is a good measure of the square of the $l_2$-norm of the overall
gradient.  Therefore, from the loss curves of Figure 1(c), it is clear that in
early training, the more the noise, the weaker the $l_2$-norm of the gradient.
If we assume that the per-example $l_2$-norm is the same in all variants at
start of training, then from Equation~\ref{eq:2}, it is clear that with greater
noise, the gradients are more dissimilar.

Finally, we note that this experiment is an instance where early stopping
(e.g., \citet{Caruana00}) is effective. Coherent gradients and the discussion
in \S\ref{sec:pred} provide some insight into this:
Strong gradients both generalize well (they are stable since they are supported
by many examples) and they bring the training loss down quickly for those
examples. 
Thus early stopping maximizes the use of strong gradients and limits the impact
of weak gradients. 
(The experiment in the \S\ref{sec:win} discusses a different way to limit the
impact of weak gradients and is an interesting point of comparison with early
stopping.)

\subsection{Analyzing Strong and Weak Gradients}
\label{sec:diff}

\begin{figure}[t]
\begin{center}
    \includegraphics[width=0.8\textwidth]{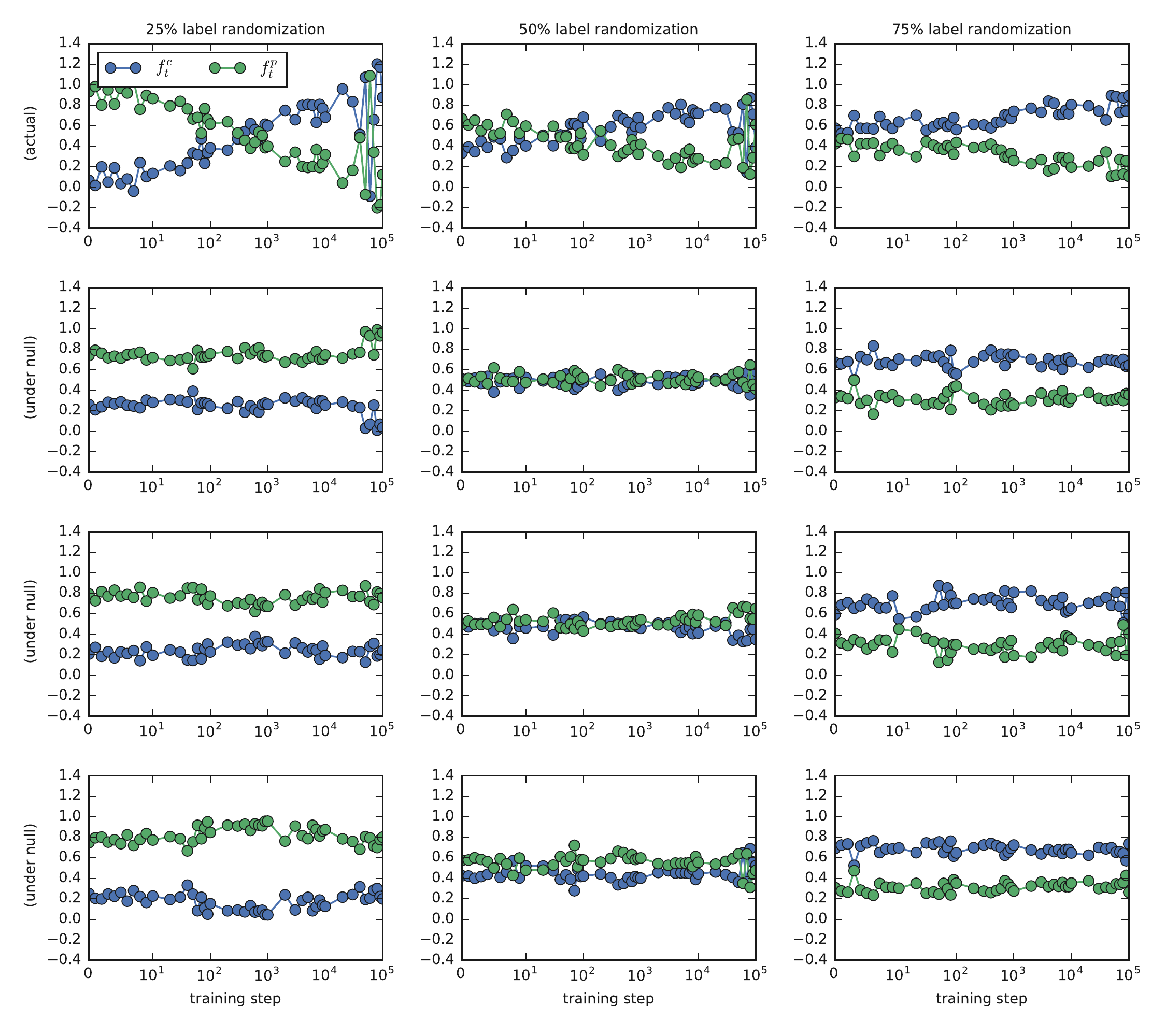}
\end{center}
    \caption{Relative contributions of pristine (similar) and corrupt (dissimilar) examples to point-in-time loss reduction. To get
    a sense of statistical significance, we show the actual statistic as well as 3 simulations under the null assuming there is
    no difference. See \S\ref{sec:diff}.}
\label{fig:fraction}
\end{figure}

Within each noisy dataset, we expect the pristine examples to be more similar
to each other and the corrupt ones to be less similar. In turn, based on the
training curves (particularly, Figure~\ref{fig:ta} (d)), during the initial
part of training, this should mean that the gradients from the pristine
examples should be stronger than the gradients from the corrupt examples. We
can study this effect via a different decomposition of square of the $l_2$-norm of the
gradient (of equivalently upto a constant, the change in the loss function):
\[
    \langle g_t, g_t \rangle 
    = \langle g_t, g_t^p + g_t^c \rangle 
    = \langle g_t,  g_t^p \rangle + \langle g_t, g_t^c \rangle
\]
where $g_t^p$ and $g_t^c$ are the sum of the gradients of the pristine examples
and corrupt examples respectively.
(We cannot decompose the overall norm into a sum of norms of pristine and
corrupt due to the cross terms $\langle g_t^p, g_t^c \rangle$. With this
decomposition, we attribute the cross terms equally to both.)
Now, set $f_t^p = \frac{\langle g_t,  g_t^p \rangle}{<g_t, g_t>}$ and $f_t^c =
\frac{ \langle g_t, g_t^c \rangle}{\langle g_t, g_t \rangle}$.
Thus, $f_t^p$ and $f_t^c$ represent the fraction of the
loss reduction due to pristine and corrupt at each time step respectively (and we have $f_t^p + f_t^c = 1$),
and based on the foregoing, we expect the pristine fraction to be
a larger fraction of the total when training starts and to diminish as training
progresses and the pristine examples are fitted.

The first row of Figure~\ref{fig:fraction} shows a plot of estimates of $f_t^p$
and $f_t^c$ for 25\%, 50\% and 75\% noise. These quantities were estimated by
recording a sample of 400 per-example gradients for 600 weights (300 from each
layer) in the network.
We see that for 25\% and 50\% label noise, $f_t^p$ initially starts off higher
than $f_t^c$ and after a few steps they cross over. This happens because at
that point all the pristine examples have been fitted and for most of the rest
of training the corrupt examples need to be fitted and so they largely contribute to
the $l_2$-norm of the gradient (or equivalently by Equation~\ref{eq:1} to loss reduction). Only at the end when the
corrupt examples have also been fit, the two curves reach parity. In the case
of 75\% noise, we see that the cross over doesn’t happen, but there is a slight
slope downwards for the contribution from pristine examples. We believe this is
because of the sheer number of corrupt examples, and so even though the
individual corrupt example gradients are weak, their sum dominates.

To get a sense of statistical significance in our hypothesis that there is a
difference between the pristine and corrupt examples as a group, in
the remaining rows of Figure~\ref{fig:fraction}, we construct a null world
where there is no difference between pristine and corrupt. We do that by
randomly permuting the ``corrupt" and ``pristine" designations among the
examples (instead of using the actual designations) and reploting. Although the
null pristine and corrupt curves are mirror images (as they must be even in the
null world since each example is given one of the two designations), we note
that for 25\% and 50\% they do not cross over as they do with the real data.
This increases our confidence that the null may be rejected. The 75\% case is
weaker but only the real data shows the slight downward slope in pristine which
none of the nulls typically show. However, all the nulls do show that corrupt
is more than pristine which increases our confidence that this is due to the
significantly differing sizes of the two sets. (Note that this happens in
reverse in the 25\% case: pristine is always above corrupt, but they never
cross over in the null worlds.)

\begin{figure}[t]
\begin{center}
    \includegraphics[width=0.8\textwidth]{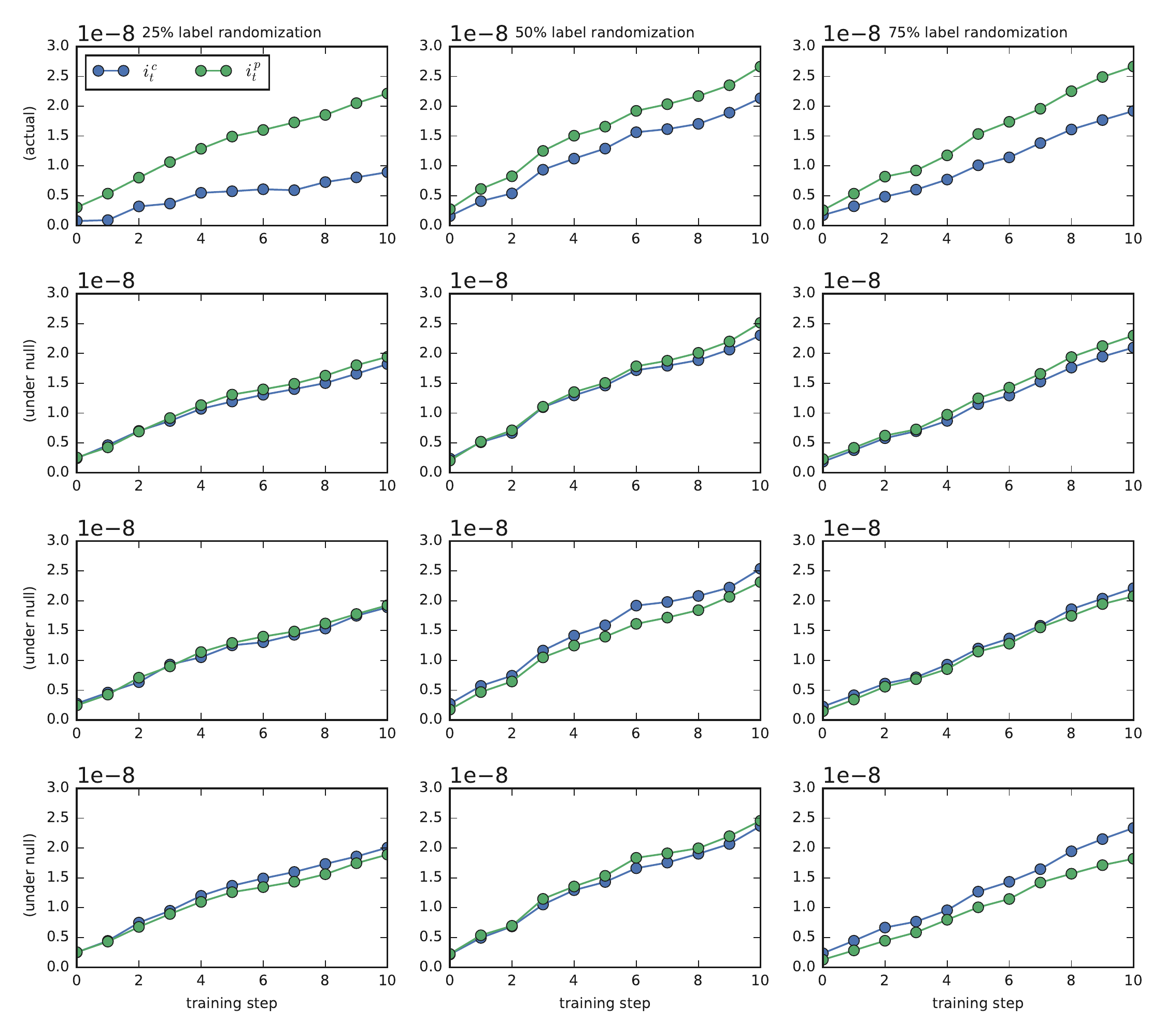}
\end{center}
    \caption{Contributions of mean pristine (similar) and corrupt (dissimilar) examples to loss reduction accumulated over first few steps of training. To get a sense of statistical significance, we show the actual statistic as well as 3 simulations under the null assuming there is no difference.
    See \S\ref{sec:diff}.}
\label{fig:cummeans}
\end{figure}

To get a stronger signal for the difference between pristine and corrupt in the
75\% case, we can look at a different statistic that adjusts for the different
sizes of the pristine and corrupt sets. Let $|p|$ and $|c|$ be the number of
pristine and corrupt examples respectively. Define
\[
    i_t^p := \frac{1}{|p|} \sum\limits_{t'=0}^{t} \langle g_{t'}, g_{t'}^p \rangle 
    \quad
    \textrm{and}
    \quad
    i_t^c := \frac{1}{|c|} \sum\limits_{t'=0}^{t} \langle g_{t'}, g_{t'}^c \rangle 
\]
which represents to a first order and upto a scale factor ($\alpha$) the
mean cumulative contribution of a pristine or corrupt example up until that
point in training (since the total change in loss from the start of training to
time $t$ is approximately the sum of first order changes in the loss at each
time step).
 
The first row of Figure~\ref{fig:cummeans} shows $i_t^p$ and $i_t^c$ for the
first 10 steps of training where the difference between pristine and corrupt is
the most pronounced. As before, to give a sense of statistical significance,
the remaining rows show the same plots in null worlds where we randomly permute
the pristine or corrupt designations of the examples. The results appear
somewhat significant but not overwhelmingly so. It would be interesting to redo
this on the entire population of examples and trainable parameters instead of
a small sample.

\section{Effect of Suppressing Weak Gradient Directions}
\label{sec:win}

In the second test of the Coherent Gradients hypothesis, we change GD itself in a very specific (and
to our knowledge, novel) manner suggested by the theory. 
Our inspiration comes from random forests. As noted in the introduction, by
building sufficiently deep trees a random forest algorithm can get perfect
training accuracy with random labels, yet generalize well when trained on real data. 
However, if we limit the tree construction algorithm to have a certain minimum
number of examples in each leaf, then it no longer overfits. 
In the case of GD, we can do something similar by suppressing the
weak gradient directions.

\subsection{Setup}

Our baseline setup is the same as before (\S\ref{sec:setup}) but we add a new dimension by modifying SGD to update each
parameter with a ``winsorized" gradient where we clip the most extreme values (outliers) among all the per-example gradients.
Formally, let $g_{we}$ be the gradient for the trainable parameter $w$ for
example $e$. The usual gradient computation for $w$ is \[ g_w = \sum\limits_e
g_{we} \]

Now let $c \in [0, 50]$ be a hyperparameter that controls the level of
winsorization. Define $l_w$ to be the $c$-th percentile of $g_{we}$ taken over
the examples. Similarly, let $u_w$ be the $(100-c)$-th percentile. Now, compute
the $c$-winsorized gradient for $w$ (denoted by $g_w^c$) as follows:
\[
    g_w^c := \sum\limits_e {\rm clip}(g_{we}, l_w, u_w)
\]
The change to gradient descent is to simply use $g_w^c$ instead of $g_w$ when
updating $w$ at each step. 
 
Note that although this is conceptually a simple change, it is computationally
very expensive due to the need for per-example gradients. To reduce the
computational cost we only use the examples in the minibatch to compute $l_w$
and $u_w$. 
Furthermore, instead of using 1 hidden layer of 2048 ReLUs, we use a smaller
network with 3 hidden layers of 256 ReLUs each, and train for 60,000 steps
(i.e., 100 epochs) with a fixed learning rate of 0.1.
We train on the baseline dataset and the 4 noisy variants with $c \in \{0, 1, 2, 4, 8\}$.
Since we have 100 examples in each minibatch, the value of $c$ immediately
tells us how many outliers are clipped in each minibatch. For example, $c=2$
means the 2 largest and 2 lowest values of the per-example gradient are
clipped (independently for each trainable parameter in the network), and $c = 0$ corresponds to unmodified SGD.

\subsection{Qualitative Predictions}

If the Coherent Gradient hypothesis is right, then the strong gradients are
responsible for making changes to the network that generalize well since they
improve many examples simultaneously. On the other hand, the weak gradients
lead to overfitting since they only improve a few examples. 
By winsorizing each coordinate, we suppress the most extreme values and thus
ensure that a parameter is only updated in a manner that benefits multiple
examples. Therefore:

\begin{itemize}

    \item Since $c$ controls which examples are considered extreme, the larger
        $c$ is, the less we expect the network to overfit. 
    
    \item But this also makes it harder for the network to fit the training
        data, and so we expect the training accuracy to fall as well.

    \item Winsorization will not completely eliminate the weak directions. For
        example, for small values of $c$ we should still expect overfitting to
        happen over time though at a reduced rate since only the most egregious
        outliers are suppressed.

\end{itemize}

\subsection{Agreement with Experiment}
\label{sec:winexpt}

\begin{figure}[t]
\begin{center}
    \includegraphics[width=\textwidth]{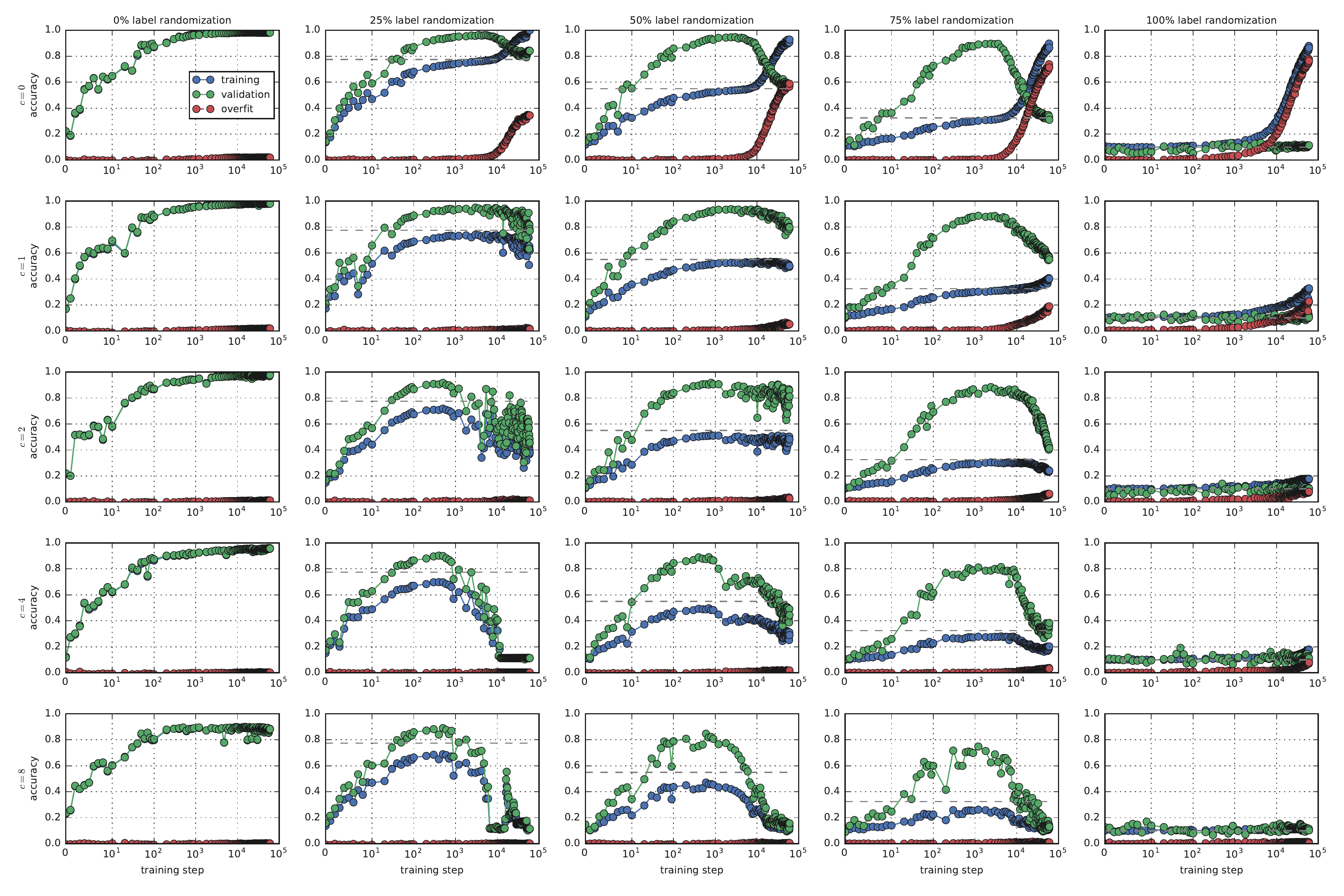}
\end{center}
\caption{Effect of suppressing weak gradient directions by eliminating outlier per-example
    gradients. This is done independently for each trainable parameter. Overfit is measured
    after accounting for the fact that test labels are not randomized (\S\ref{sec:winexpt}).}
\label{fig:winsor}
\vskip -0.2in
\end{figure}

The resulting training and test curves shown in Figure~\ref{fig:winsor}. The
columns correspond to different amounts of label noise and the rows to
different amounts of winsorization.
In addition to the training and test accuracies (${\sf ta}$ and ${\sf va}$,
respectively), we show the level of overfit which is defined as ${\sf ta} - [
    \epsilon \cdot \frac{1}{10}  + (1 - \epsilon) \cdot {\sf va}]$ to account
for the fact that the test labels are not randomized.

We see that the experimental results are in agreement with the predictions above. In particular,

\begin{itemize}

    \item For $c > 1$, training accuracies do not exceed the proper accuracy of the dataset, though they may fall short, specially for large values of $c$.

    \item The rate at which the overfit curve grows goes down with increasing $c$.

\end{itemize}

Additionally, we notice that with a large amount of winsorization, the training and
test accuracies reach a maximum and then go down. Part of the reason is that as
a result of winsorization, each step is no longer in a descent direction, i.e.,
this is no longer gradient descent.

\section{Discussion and Related Work}
\label{sec:related}

Although there has been a lot of work in recent years in trying to understand
generalization in Deep Learning, no entirely satisfactory explanation has
emerged so far. 

There is a rich literature on aspects of the stochastic optimization problem
such as the loss landscape and minima
(e.g., \citet{Choromanska15, Zhu18}), the curvature around stationary points
(e.g., \citet{Hochreiter97, Keskar16, Dinh17, Wu18}), and the implications of
stochasticity due to sampling in SGD (e.g., \citet{Simsekli19}).
However, we believe it should be possible to understand generalization without
a detailed understanding of the optimization landscape.  For example, since
stopping early typically leads to small generalization gap, the nature of the
solutions of GD (e.g., stationary points, the limit cycles of SGD at
equilibrium) cannot be solely responsible for generalization. 
In fact, from this observation, it would appear that an inductive argument for
generalization would be more natural.  
Likewise, there is reason to believe that stochasticity is not fundamental to
generalization (though it may help). For example, modifying the experiment in
\S\ref{sec:setup} to use full batch leads to similar qualitative generalization
results.  This is consistent with other small scale studies (e.g., Figure 1
of~\citet{Wu18}) though we are not aware of any large scale studies on full
batch.

Our view of optimization is a simple, almost combinatorial, one: gradient
descent is a greedy search with some hill-climbing thrown in (due to sampling
in SGD and finite step size). Therefore, we worry less about the quality of
solutions reached, but more about staying ``feasible" at all times during the
search. In our context, feasibility means being able to generalize; and this
naturally leads us to look at the transition dynamics to see if that preserves
generalizability.

Another approach to understanding generalization, is to argue that gradient-based
optimization induces a form of implicit regularization leading to a bias towards models
of low complexity.
This is an extension of the classical approach 
where bounding a complexity measure leads to bounds on the generalization
gap. As is well known, classical measures of complexity (also called capacity)
do not work well. 
For example, sometimes adding more parameters to a
net can help generalization (see for e.g.~\citet{Lawrence96,Neyshabur18}) and,
as we have seen, VC-Dimension and Rademacher Complexity-based bounds must be
vacuous since networks can memorize random labels and yet generalize on real
data. 
This has led to a lot of recent work in identifying better measures of
complexity such as spectrally-normalized margin~\citep{Bartlett17}, path-based
group norm~\citep{Neyshabur18}, a compression-based approach~\citep{Arora18},
etc. However, to our knowledge, none of these measures is entirely satisfactory
for accounting for generalization in practice. Please see~\citet{Nagarajan19} for
an excellent discussion of the challenges. 

We rely on a different classical notion to argue generalization: algorithmic
stability (see \citet{Bousquet02} for a historical overview). We have provided
only an informal argument in Section 1, but there has been prior work by
\citet{Hardt16} in looking at GD and SGD through the lens of stability, but
their formal results do not explain generalization in practical settings (e.g.,
multiple epochs of training and non-convex objectives). In fact, such an
attempt appears unlikely to work since our experimental results imply that any
stability bounds for SGD that do not account for the actual training data must
be vacuous! (This was also noted by \citet{Zhang17}.)
That said, we believe stability is the right way to think about generalization
in GD for a few reasons. First, since by \citet{ShalevShwartz10} stability,
suitably formalized, is equivalent to generalization. Therefore, in principle,
any explanation of generalizability for a learning problem must---to borrow a
term from category theory---factor through stability. Second, a stability based
analysis may be more amenable to taking the actual training data into account
(perhaps by using a ``stability accountant" similar to a privacy accountant)
which appears necessary to get non-vacuous bounds for practical networks and
datasets. Finally, as we have seen with the modification in \S\ref{sec:win}, a
stability based approach is not just descriptive but prescriptive\footnote{See
https://www.offconvex.org/2017/12/08/generalization1/ for a nice discussion of
the difference.} and can point the way to better learning algorithms.

Finally, we look at two relevant lines of work pointed out by a reviewer.
First, \citet{Rahaman19} compute the Fourier spectrum
of ReLU networks and argue based on heuristics and experiments that these
networks learn low frequency functions first. In contrast, we focus not on the
function learnt, but on the mechanism in GD to detect commonality. This leads
to a perspective that is at once simpler and more general (for e.g., it applies
equally to networks with other activation functions, with attention, LSTMs, and discrete
(combinatorial) inputs). Furthermore, it opens up a path to analyzing
generalization via stability.
It is is not clear if \citet{Rahaman19} claim a causal mechanism, but their
analysis does not suggest an obvious intervention experiment such as ours of
\S\ref{sec:win} to test causality.
There are other experimental results that show biases towards linear
functions~\citep{Nakkiran19} and functions with low descriptive
complexity~\citep{VallePerez19} but these papers do not posit a causal
mechanism. It is interesting to consider if Coherent Gradients can provide a
unified explanation for these observed biases.
 
Second, \citet{Fort19} propose a descriptive statistic
{\em stiffness} based on pairwise per-example gradients and show experimentally
that it can be used to characterize generalization. \citet{Sankararaman19}
propose a very similar statistic
called {\em gradient confusion} but use it to study the speed of training.
Unlike our work, these do not propose causal mechanisms for generalization, but
these statistics (which are different from those in \S\ref{sec:diff})
could be useful for the further study of Coherent Gradients.

\section{Directions for Future Work}
\label{sec:future}

Does the Coherent Gradients hypothesis hold in other settings such as BERT,
ResNet, etc.? For that we would need to develop more computationally efficient
tests. Can we use the state of the network to explicitly characterize which
examples are considered similar and study this evolution in the course of
training?
We expect non-parametric methods for similarity such as those developed in
\citet{Chatterjee19} and their characterization of ``easy" examples (i.e.,
examples learnt early as per \citet{Arpit17}) as those with many others like
them, to be useful in this context.

Can Coherent Gradients explain adversarial initializations~\citep{Liu19}?  The
adversarial initial state makes semantically similar examples purposefully look
different. Therefore, during training, they continue to be treated differently
(i.e., their gradients share less in common than they would if starting from a
random initialization). Thus, fitting is more case-by-case and while it
achieves good final training accuracy, it does not generalize.

Can Coherent Gradients along with the Lottery Ticket
Hypothesis~\citep{Frankle18} explain the observation in \citet{Neyshabur18}
that wider networks generalize better? By Lottery Ticket, wider networks
provide more chances to find initial gradient directions that improve many
examples, and by Coherent Gradients, these popular hypotheses are learned
preferentially (faster).

Can we use the ideas behind Winsorized SGD from \S\ref{sec:win} to develop a
computationally efficient learning algorithm with generalization (and even
privacy) guarantees?
How does winsorized gradients compare in practice to the algorithm proposed
in~\cite{Abadi16} for privacy? Last, but not least, can we use the insights
from this work to design learning algorithms that operate natively on discrete
networks?


\subsubsection*{Acknowledgments}
I thank Alan Mishchenko, Shankar Krishnan, Piotr Zielinski,
Chandramouli Kashyap, Sergey Ioffe, Michele Covell, and Jay Yagnik for helpful
discussions.

\bibliography{paper}

\begin{thebibliography}{30}
\providecommand{\natexlab}[1]{#1}
\providecommand{\url}[1]{\texttt{#1}}
\expandafter\ifx\csname urlstyle\endcsname\relax
  \providecommand{\doi}[1]{doi: #1}\else
  \providecommand{\doi}{doi: \begingroup \urlstyle{rm}\Url}\fi

\bibitem[Abadi et~al.(2016)Abadi, Chu, Goodfellow, McMahan, Mironov, Talwar,
  and Zhang]{Abadi16}
Martin Abadi, Andy Chu, Ian Goodfellow, H.~Brendan McMahan, Ilya Mironov, Kunal
  Talwar, and Li~Zhang.
\newblock Deep learning with differential privacy.
\newblock In \emph{Proceedings of the 2016 ACM SIGSAC Conference on Computer
  and Communications Security}, CCS '16, pp.\  308--318, New York, NY, USA,
  2016. ACM.
\newblock ISBN 978-1-4503-4139-4.
\newblock \doi{10.1145/2976749.2978318}.
\newblock URL \url{http://doi.acm.org/10.1145/2976749.2978318}.

\bibitem[Arora et~al.(2018)Arora, Ge, Neyshabur, and Zhang]{Arora18}
Sanjeev Arora, Rong Ge, Behnam Neyshabur, and Yi~Zhang.
\newblock Stronger generalization bounds for deep nets via a compression
  approach.
\newblock In Jennifer~G. Dy and Andreas Krause (eds.), \emph{Proceedings of the
  35th International Conference on Machine Learning, {ICML} 2018,
  Stockholmsm{\"{a}}ssan, Stockholm, Sweden, July 10-15, 2018}, volume~80 of
  \emph{Proceedings of Machine Learning Research}, pp.\  254--263. {PMLR},
  2018.
\newblock URL \url{http://proceedings.mlr.press/v80/arora18b.html}.

\bibitem[Arpit et~al.(2017)Arpit, Jastrzebski, Ballas, Krueger, Bengio, Kanwal,
  Maharaj, Fischer, Courville, Bengio, and Lacoste{-}Julien]{Arpit17}
Devansh Arpit, Stanislaw~K. Jastrzebski, Nicolas Ballas, David Krueger,
  Emmanuel Bengio, Maxinder~S. Kanwal, Tegan Maharaj, Asja Fischer, Aaron~C.
  Courville, Yoshua Bengio, and Simon Lacoste{-}Julien.
\newblock A closer look at memorization in deep networks.
\newblock In \emph{Proceedings of the 34th International Conference on Machine
  Learning, {ICML} 2017, Sydney, NSW, Australia, 6-11 August 2017}, pp.\
  233--242, 2017.
\newblock URL \url{http://proceedings.mlr.press/v70/arpit17a.html}.

\bibitem[Bartlett et~al.(2017)Bartlett, Foster, and Telgarsky]{Bartlett17}
Peter~L Bartlett, Dylan~J Foster, and Matus~J Telgarsky.
\newblock Spectrally-normalized margin bounds for neural networks.
\newblock In I.~Guyon, U.~V. Luxburg, S.~Bengio, H.~Wallach, R.~Fergus,
  S.~Vishwanathan, and R.~Garnett (eds.), \emph{Advances in Neural Information
  Processing Systems 30}, pp.\  6240--6249. Curran Associates, Inc., 2017.

\bibitem[Belkin et~al.(2019)Belkin, Hsu, Ma, and Mandal]{Belkin19}
Mikhail Belkin, Daniel Hsu, Siyuan Ma, and Soumik Mandal.
\newblock Reconciling modern machine-learning practice and the classical
  bias{\textendash}variance trade-off.
\newblock \emph{Proceedings of the National Academy of Sciences}, 116\penalty0
  (32):\penalty0 15849--15854, 2019.
\newblock ISSN 0027-8424.
\newblock \doi{10.1073/pnas.1903070116}.
\newblock URL \url{https://www.pnas.org/content/116/32/15849}.

\bibitem[Bousquet \& Elisseeff(2002)Bousquet and Elisseeff]{Bousquet02}
Olivier Bousquet and Andr{\'e} Elisseeff.
\newblock Stability and generalization.
\newblock \emph{J. Mach. Learn. Res.}, 2:\penalty0 499--526, March 2002.
\newblock ISSN 1532-4435.
\newblock \doi{10.1162/153244302760200704}.
\newblock URL \url{https://doi.org/10.1162/153244302760200704}.

\bibitem[Caruana et~al.(2000)Caruana, Lawrence, and Giles]{Caruana00}
Rich Caruana, Steve Lawrence, and Lee Giles.
\newblock Overfitting in neural nets: Backpropagation, conjugate gradient, and
  early stopping.
\newblock In \emph{Proceedings of the 13th International Conference on Neural
  Information Processing Systems}, NIPS'00, pp.\  381--387, Cambridge, MA, USA,
  2000. MIT Press.
\newblock URL \url{http://dl.acm.org/citation.cfm?id=3008751.3008807}.

\bibitem[Chatterjee \& Mishchenko(2019)Chatterjee and Mishchenko]{Chatterjee19}
Satrajit Chatterjee and Alan Mishchenko.
\newblock Circuit-based intrinsic methods to detect overfitting.
\newblock \emph{CoRR}, abs/1907.01991, 2019.
\newblock URL \url{http://arxiv.org/abs/1907.01991}.

\bibitem[Choromanska et~al.(2015)Choromanska, Henaff, Mathieu, Arous, and
  LeCun]{Choromanska15}
Anna Choromanska, Mikael Henaff, Micha{\"{e}}l Mathieu, G{\'{e}}rard~Ben Arous,
  and Yann LeCun.
\newblock The loss surfaces of multilayer networks.
\newblock In Guy Lebanon and S.~V.~N. Vishwanathan (eds.), \emph{Proceedings of
  the Eighteenth International Conference on Artificial Intelligence and
  Statistics, {AISTATS} 2015, San Diego, California, USA, May 9-12, 2015},
  volume~38 of \emph{{JMLR} Workshop and Conference Proceedings}. JMLR.org,
  2015.
\newblock URL \url{http://proceedings.mlr.press/v38/choromanska15.html}.

\bibitem[Dinh et~al.(2017)Dinh, Pascanu, Bengio, and Bengio]{Dinh17}
Laurent Dinh, Razvan Pascanu, Samy Bengio, and Yoshua Bengio.
\newblock Sharp minima can generalize for deep nets.
\newblock \emph{CoRR}, abs/1703.04933, 2017.
\newblock URL \url{http://arxiv.org/abs/1703.04933}.

\bibitem[Fort et~al.(2019)Fort, Nowak, and Narayanan]{Fort19}
Stanislav Fort, Pawel~Krzysztof Nowak, and Srini Narayanan.
\newblock Stiffness: {A} new perspective on generalization in neural networks.
\newblock \emph{CoRR}, abs/1901.09491, 2019.
\newblock URL \url{http://arxiv.org/abs/1901.09491}.

\bibitem[Frankle \& Carbin(2018)Frankle and Carbin]{Frankle18}
Jonathan Frankle and Michael Carbin.
\newblock The lottery ticket hypothesis: Training pruned neural networks.
\newblock \emph{CoRR}, abs/1803.03635, 2018.
\newblock URL \url{http://arxiv.org/abs/1803.03635}.

\bibitem[Hardt et~al.(2016)Hardt, Recht, and Singer]{Hardt16}
Moritz Hardt, Benjamin Recht, and Yoram Singer.
\newblock Train faster, generalize better: Stability of stochastic gradient
  descent.
\newblock In \emph{Proceedings of the 33rd International Conference on
  International Conference on Machine Learning - Volume 48}, ICML'16, pp.\
  1225--1234. JMLR.org, 2016.
\newblock URL \url{http://dl.acm.org/citation.cfm?id=3045390.3045520}.

\bibitem[Hochreiter \& Schmidhuber(1997)Hochreiter and
  Schmidhuber]{Hochreiter97}
Sepp Hochreiter and J\"{u}rgen Schmidhuber.
\newblock Flat minima.
\newblock \emph{Neural Comput.}, 9\penalty0 (1):\penalty0 1--42, January 1997.
\newblock ISSN 0899-7667.
\newblock \doi{10.1162/neco.1997.9.1.1}.
\newblock URL \url{http://dx.doi.org/10.1162/neco.1997.9.1.1}.

\bibitem[{Kawaguchi} et~al.(2017){Kawaguchi}, {Pack Kaelbling}, and
  {Bengio}]{Kawaguchi17}
K.~{Kawaguchi}, L.~{Pack Kaelbling}, and Y.~{Bengio}.
\newblock {Generalization in Deep Learning}.
\newblock \emph{ArXiv e-prints}, December 2017.
\newblock URL \url{https://arxiv.org/abs/1710.05468v2}.

\bibitem[Keskar et~al.(2016)Keskar, Mudigere, Nocedal, Smelyanskiy, and
  Tang]{Keskar16}
Nitish~Shirish Keskar, Dheevatsa Mudigere, Jorge Nocedal, Mikhail Smelyanskiy,
  and Ping Tak~Peter Tang.
\newblock On large-batch training for deep learning: Generalization gap and
  sharp minima.
\newblock \emph{CoRR}, abs/1609.04836, 2016.
\newblock URL \url{http://arxiv.org/abs/1609.04836}.

\bibitem[Lawrence et~al.(1996)Lawrence, Giles, and Tsoi]{Lawrence96}
Steve Lawrence, C.~Lee Giles, and Ah~Chung Tsoi.
\newblock What size neural network gives optimal generalization? convergence
  properties of backpropagation.
\newblock Technical report, 1996.

\bibitem[Liu et~al.(2019)Liu, Papailiopoulos, and Achlioptas]{Liu19}
Shengchao Liu, Dimitris~S. Papailiopoulos, and Dimitris Achlioptas.
\newblock Bad global minima exist and {SGD} can reach them.
\newblock \emph{CoRR}, abs/1906.02613, 2019.
\newblock URL \url{http://arxiv.org/abs/1906.02613}.

\bibitem[Nagarajan \& Kolter(2019)Nagarajan and Kolter]{Nagarajan19}
Vaishnavh Nagarajan and J.~Zico Kolter.
\newblock Uniform convergence may be unable to explain generalization in deep
  learning.
\newblock In Hanna~M. Wallach, Hugo Larochelle, Alina Beygelzimer, Florence
  d'Alch{\'{e}}{-}Buc, Emily~B. Fox, and Roman Garnett (eds.), \emph{Advances
  in Neural Information Processing Systems 32: Annual Conference on Neural
  Information Processing Systems 2019, NeurIPS 2019, 8-14 December 2019,
  Vancouver, BC, Canada}, pp.\  11611--11622, 2019.

\bibitem[Nakkiran et~al.(2019)Nakkiran, Kaplun, Kalimeris, Yang, Edelman,
  Zhang, and Barak]{Nakkiran19}
Preetum Nakkiran, Gal Kaplun, Dimitris Kalimeris, Tristan Yang, Benjamin~L.
  Edelman, Fred Zhang, and Boaz Barak.
\newblock {SGD} on neural networks learns functions of increasing complexity.
\newblock \emph{CoRR}, abs/1905.11604, 2019.
\newblock URL \url{http://arxiv.org/abs/1905.11604}.

\bibitem[Neyshabur et~al.(2018)Neyshabur, Li, Bhojanapalli, LeCun, and
  Srebro]{Neyshabur18}
Behnam Neyshabur, Zhiyuan Li, Srinadh Bhojanapalli, Yann LeCun, and Nathan
  Srebro.
\newblock Towards understanding the role of over-parametrization in
  generalization of neural networks.
\newblock \emph{CoRR}, abs/1805.12076, 2018.
\newblock URL \url{http://arxiv.org/abs/1805.12076}.

\bibitem[Rahaman et~al.(2019)Rahaman, Baratin, Arpit, Draxler, Lin, Hamprecht,
  Bengio, and Courville]{Rahaman19}
Nasim Rahaman, Aristide Baratin, Devansh Arpit, Felix Draxler, Min Lin, Fred
  Hamprecht, Yoshua Bengio, and Aaron Courville.
\newblock On the spectral bias of neural networks.
\newblock In Kamalika Chaudhuri and Ruslan Salakhutdinov (eds.),
  \emph{Proceedings of the 36th International Conference on Machine Learning},
  volume~97 of \emph{Proceedings of Machine Learning Research}, pp.\
  5301--5310, Long Beach, California, USA, 09--15 Jun 2019. PMLR.
\newblock URL \url{http://proceedings.mlr.press/v97/rahaman19a.html}.

\bibitem[Rolnick et~al.(2017)Rolnick, Veit, Belongie, and Shavit]{Rolnick17}
David Rolnick, Andreas Veit, Serge~J. Belongie, and Nir Shavit.
\newblock Deep learning is robust to massive label noise.
\newblock \emph{CoRR}, abs/1705.10694, 2017.
\newblock URL \url{http://arxiv.org/abs/1705.10694}.

\bibitem[Sankararaman et~al.(2019)Sankararaman, De, Xu, Huang, and
  Goldstein]{Sankararaman19}
Karthik~Abinav Sankararaman, Soham De, Zheng Xu, W.~Ronny Huang, and Tom
  Goldstein.
\newblock The impact of neural network overparameterization on gradient
  confusion and stochastic gradient descent.
\newblock \emph{CoRR}, abs/1904.06963, 2019.
\newblock URL \url{http://arxiv.org/abs/1904.06963}.

\bibitem[Shalev-Shwartz et~al.(2010)Shalev-Shwartz, Shamir, Srebro, and
  Sridharan]{ShalevShwartz10}
Shai Shalev-Shwartz, Ohad Shamir, Nathan Srebro, and Karthik Sridharan.
\newblock Learnability, stability and uniform convergence.
\newblock \emph{J. Mach. Learn. Res.}, 11:\penalty0 2635--2670, December 2010.
\newblock ISSN 1532-4435.
\newblock URL \url{http://dl.acm.org/citation.cfm?id=1756006.1953019}.

\bibitem[Simsekli et~al.(2019)Simsekli, Sagun, and
  G{\"{u}}rb{\"{u}}zbalaban]{Simsekli19}
Umut Simsekli, Levent Sagun, and Mert G{\"{u}}rb{\"{u}}zbalaban.
\newblock A tail-index analysis of stochastic gradient noise in deep neural
  networks.
\newblock In Kamalika Chaudhuri and Ruslan Salakhutdinov (eds.),
  \emph{Proceedings of the 36th International Conference on Machine Learning,
  {ICML} 2019, 9-15 June 2019, Long Beach, California, {USA}}, volume~97 of
  \emph{Proceedings of Machine Learning Research}, pp.\  5827--5837. {PMLR},
  2019.
\newblock URL \url{http://proceedings.mlr.press/v97/simsekli19a.html}.

\bibitem[Valle-Perez et~al.(2019)Valle-Perez, Camargo, and Louis]{VallePerez19}
Guillermo Valle-Perez, Chico~Q. Camargo, and Ard~A. Louis.
\newblock Deep learning generalizes because the parameter-function map is
  biased towards simple functions.
\newblock In \emph{International Conference on Learning Representations}, 2019.
\newblock URL \url{https://openreview.net/forum?id=rye4g3AqFm}.

\bibitem[Wu et~al.(2018)Wu, Ma, and E]{Wu18}
Lei Wu, Chao Ma, and Weinan E.
\newblock How sgd selects the global minima in over-parameterized learning: A
  dynamical stability perspective.
\newblock In S.~Bengio, H.~Wallach, H.~Larochelle, K.~Grauman, N.~Cesa-Bianchi,
  and R.~Garnett (eds.), \emph{Advances in Neural Information Processing
  Systems 31}, pp.\  8279--8288. Curran Associates, Inc., 2018.

\bibitem[Zhang et~al.(2017)Zhang, Bengio, Hardt, Recht, and Vinyals]{Zhang17}
Chiyuan Zhang, Samy Bengio, Moritz Hardt, Benjamin Recht, and Oriol Vinyals.
\newblock Understanding deep learning requires rethinking generalization.
\newblock In \emph{Proceedings of the International Conference on Learning
  Representations {ICLR}}, 2017.

\bibitem[Zhu et~al.(2018)Zhu, Soudry, Eldar, and Wakin]{Zhu18}
Zhihui Zhu, Daniel Soudry, Yonina~C. Eldar, and Michael~B. Wakin.
\newblock The global optimization geometry of shallow linear neural networks.
\newblock \emph{CoRR}, abs/1805.04938, 2018.
\newblock URL \url{http://arxiv.org/abs/1805.04938}.

\end{thebibliography}
\bibliographystyle{iclr2020_conference}


\end{document}